\documentclass{iopconfser}

\usepackage{xcolor}
\usepackage{hyperref}
\hypersetup{
    colorlinks,
    linkcolor={red!50!black},
    citecolor={blue!50!black},
    urlcolor={blue!80!black}
}

\usepackage{amsmath}
\usepackage[numbers,sort&compress]{natbib}
\usepackage{lineno}
\usepackage[bitstream-charter]{mathdesign}
\DeclareSymbolFont{usualmathcal}{OMS}{cmsy}{m}{n}
\DeclareSymbolFontAlphabet{\mathcal}{usualmathcal}

\usepackage{graphicx}
\graphicspath{{figures/}}

\usepackage{xspace}
\newcommand{\modelname}{\textsc{TrackSorter}\xspace}
\newcommand{\pt}{\ensuremath{p_\text{T}}\xspace}

\newcommand{\sep}{[SEP]\xspace}

\begin{document}

\title{\modelname: A Transformer-based sorting algorithm for track finding in High Energy Physics}

\author{Yash Melkani$^1$ and Xiangyang Ju$^2$}
\affil{$^1$University of California-Berkeley}
\affil{$^2$Lawrence Berkeley National Laboratory}
\email{yashmelkani02@gmail.com, xju@lbl.gov}

\begin{abstract}
Track finding in particle data is a challenging pattern recognition problem in High Energy Physics. It takes as inputs a point cloud of space points and labels them so that space points created by the same particle have the same label. The list of space points with the same label is a track candidate. We argue that this pattern recognition problem can be formulated as a sorting problem of which the inputs are a list of space points sorted by their distances away from the collision points and the outputs are the  space points sorted by their labels. In this paper, we propose the \modelname algorithm: a Transformer-based algorithm for  pattern recognition in particle data. \modelname uses a simple tokenization scheme to convert space points into discrete tokens. It then uses the tokenized space points as inputs and sorts the input tokens into track candidates. \modelname is a novel end-to-end track finding algorithm that leverages Transformer-based models to solve pattern recognition problems. It is evaluated on the TrackML dataset and has good track finding performance.
\end{abstract}

\section{Introduction}
\label{sec:intro}
The High Luminosity Large Hadronic Collider (HL-LHC) plans to collide two proton beams at the unprecedented center of mass energy of 14 TeV at an instantaneous luminosity of up to $7.5\times10^{34}\text{cm}^{-2}\text{s}^{-1}$. That corresponds to an average number of proton-proton collisions per beam crossing (i.e. pileup), $\langle\mu\rangle$, of up to 200. HL-LHC brings opportunities and challenges. To cope with the challenges, the ATLAS~\cite{ATLAS:2008xda} and CMS~\cite{CMS:2008xjf} experiments decided to build a new fully silicon-based inner tracker detector~\cite{ATLAS-TDR-25,ATLAS-TDR-30,CMS:2017lum}. The new inner trackers will have better raditation tolerance, increased granularity, reduced material, and large readout bandwith to fulfill the requirement of the HL-LHC Runs. Take the ATLAS's new innder detector, ITk, as an example. ITk consists of a Pixel detector at a small radius and a large area Strip detector surrounding it. The Pixel detector consists of about five billion finely segmented silicon sensors, most of which have a pitch of $50\times50\,\mu\text{m}^2$ and the rest $25\times100\,\mu\text{m}^2$. The Strip detector comprises 23,000 long and skinny silicon sensors ($75.5\,\mu\text{m}\times 24.1$ or 48.2 mm). Each event with $\langle\mu\rangle=200$ produces about 300,000 space points, out of them only about 10,000 space points come from particles of interest. Finding the tracks of interests from a point cloud of space points is a challenging pattern recognition problem. 

Our work is inspired by the remarkable capabilities of Large Language Models (LLMs) such as BERT~\cite{bert}, GPT~\cite{gpt3}, Llama~\cite{llama:code}, and grok-1~\cite{grok1}. The foundation of our work is to convert space points into discrete token ids (i.e. tokenization). Tokenization is a critical step in efficient learning and handling out-of-vocabulary words for LLMs learning natural language. For example, one can use letters as tokens for the English language. Doing so would result in a small vocabulary and can construct all out-of-vocabulary words. However, it is inefficient for learning because semantic relationships among letters are lost during tokenization. 
Tokenizing a point cloud of measurements in High Energy Physics (HEP) presents a unique challenge: converting variables from a continuous, multi-dimensional space into discrete spaces. Although some information from the continuous space will inevitably be lost during tokenization, this loss may be acceptable as long as it does not compromise the accuracy of the underlying physics. After all, all physics measurements inherently contain some level of uncertainties.
In the context of jet physics, the Omnijet framework ~\cite{Birk:2024knn} explored three schemes for tokenizing jet constituents: physics-inspired binning of contitunents' kinematic variables (\textit{binning} in short), conditional and unconditional tokenization via the vector-quantization variational autoencoder (i.e. VQ-VAE)~\cite{vqvae} technique, which is also used in Ref~\cite{Heinrich:2024sbg} to build the codebook index. The \textit{binning} scheme adjusts the bin width to match measurement uncertainties and the study shows a small bias in the reconstructed jet mass and poor resolution (Fig. 4 in Ref.~\cite{Heinrich:2024sbg}). The VQ-VAE schemes enjoy better performances with a larger number of tokens. In our previous research~\cite{Huang:2024voo} on particle tracking data, we utilized the unique detector module IDs as the token identifiers for space points created on that detector module. While this tokenization process results in the loss of precise positional and other details of the space points, it enables us to directly train LLMs with the tracking data.

\begin{figure}[!htb]
    \centering
    \includegraphics[width=0.9\textwidth]{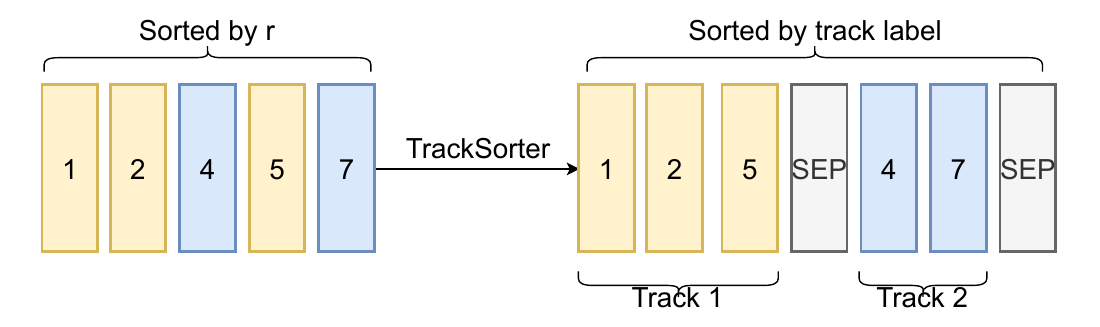}
    \caption{Illustration of the \modelname algorithm. Each box represents a space point, with the token ID inside. \sep is a special token indicating the end of a track. $r$ is the distance between the space point and the collision point in the transverse plan.}
    \label{fig:main-approach}
\end{figure}

We argue that the track finding problem can be formulated as a sequence-to-sequence (seq2seq) problem, illustrated in Fig.~\ref{fig:main-approach}. The input sequence is a list of space points ordered by their distances away from the collision point. And the output sequence is a list of the same space points ordered by their labels and their distances away from the collision point. Language models like Bart~\cite{bart} are very good at solving seq2seq problems like machine translation, text summary, and question-answering. The Bart model consists of a bidirectional encoder (like BERT) and a left-to-right decoder (like GPT). A similar model is used in our work.

\section{Data}
\label{sec:data}
\begin{figure}[!htb]
    \centering
    \includegraphics[width=0.9\textwidth]{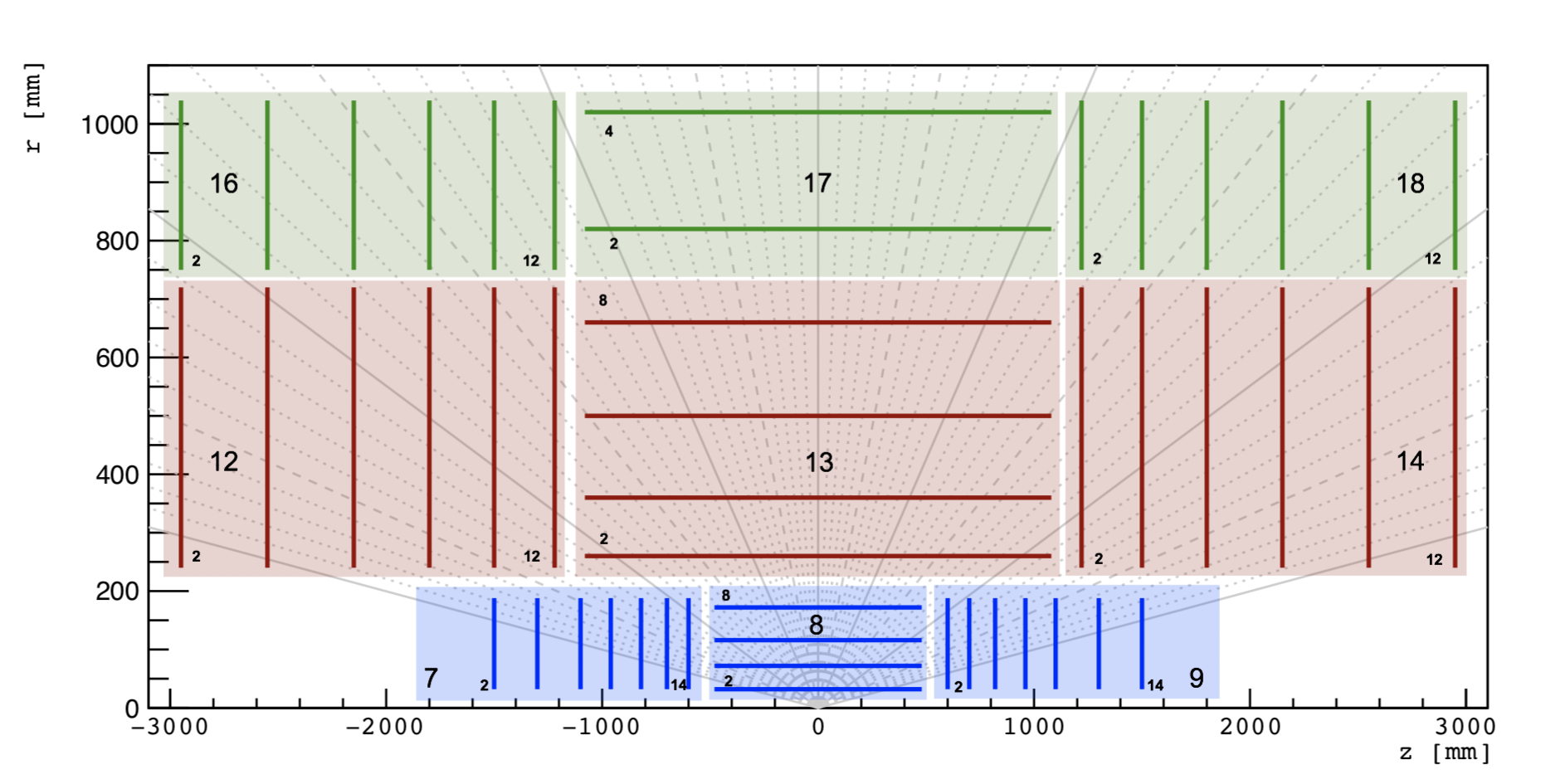}
    \caption{The detector schematic shows the top half of the detector projected on the r-z plane. The z-axis is along the beam direction.}
        \label{fig:trackml_detector}
\end{figure}

This study is based on the TrackML dataset~\cite{TrackMLAccuracy2019}, which simulates the top quark pair production from proton-proton collisions at the HL-LHC. To simulate the effect of event pileup and produce realistic detector occupancy, a Poisson random number (with $\mu=200$) of Quantum Chromodynamics "minimum bias" events are overlaid on top of the $t\bar{t}$ collisions. The TrackML detector is a set of concentric cylindrical layers of pixelated sensors (the \textit{barrel}) complemented by a set of circular disks (the \textit{endcaps}) to ensure nearly $4\pi$ coverage in solid angle, as pictured in Fig.~\ref{fig:trackml_detector}.

The detector is divided into nine volumes, each consisting of 2 to 7 layers. Each layer contains multiple silicon modules. There are 18,737 detector modules in the TrackML dataset. We use data from volume 8, 13, and 17, summing up to 14,000 modules. We introduce two custom tokens to indicate the start of the output sequence \textsc{[SOS]} and the end of each track \sep. Therefore, as detector module IDs are treated as token identifiers, the vocabulary size in our work is the sum of the number of detector modules and the two special tokens; that's 14,002.  

Our study utilizes particles that have space points from at least 6 unique layers. Our training dataset uses tracks from 100 events that meet this condition, totaling 349k tracks. The validation dataset is similarly constructed from 10 events (35k tracks). A testing dataset for performance analysis is curated with an additional condition that each track has an average $\pt < 5$ GeV, containing 67k tracks.

In natural language processing workflows, discrete tokens are first embedded into a continuous, dense vector representation, i.e. Word2Vec~\cite{Mikolov:2013ohu,word2vec}. We used the continuous bag of words framework~\cite{Mikolov:2013ohu} to train a Word2Vec model by asking the model to predict a target word using all the words in a context window. To train the model, we randomly pair each track with another track and construct a target sentence following the ordering scheme shown in Fig.~\ref{fig:main-approach} for each track pair. In total, our training data contains 349k sentences and 8M tokens. The ``Generate Similar` (Gensim) library~\cite{rehurek_lrec} is utilized for training. To achieve a reasonable embedding performance, the model uses an embedding dimension of 64, a context window of 20 tokens, and is trained for 100 epochs. In the future, we can use the TrackingBERT~\cite{Huang:2024voo} method to embed detector modules.

\section{Model and Training}
\label{sec:model}
The model utilizes the encoder-decoder structure of transformers~\cite{transformer}. Both encoder and decoder networks are composed of a stack of identical transformer modules, each having a single-head self-attention mechanism and a position-wise fully connected feed-forward network. We only use a single attention head because our embedding dimension is 64, which is relatively small. And the feed forward layers have a dimension of 256. The output of the decoder network is fed into a linear layer that spans the dimension of the vocabulary. Our model contains six bi-directional encoder layers followed by six left-to-right decoder layers, totaling 1.6M trainable parameters. We inject positional encoding into the input sequence to provide information on the order of the detector modules. 

The model is trained to autoregressively predict the correct sequence of tokens for 371 epochs using the Adam optimizer~\cite{adam} in conjunction with the CosineAnnealingLR scheduler. Model weights corresponding to the lowest validation loss were saved.

\section{Results and Discussions}
\label{sec:results}
During model inference, we utilize the greedy search algorithm to construct model predictions: given an input sequence, a count mask is created to store the number of instances of each token in the input sequence. The count value for the [SOS] and \sep tokens are set to 0 and 100, respectively. The model is first fed with the [SOS] token and predicts the next token by calculating logits for each token in the vocabulary. The logits of tokens that have a value of zero in the count mask are set to zero. The token with the highest model logit is considered as the next token; thus, it is appended to the output sequence. Its corresponding count value is decremented by one so that it would not appear again in the output sequence. The updated output sequence is then fed back to the model for the next token prediction. In the case that the greedy algorithm predicts the \sep token, the model logit of the \sep token will be set to zero in the next step. 
This is to prevent the algorithm from predicting the \sep token in consecutive steps. 
Note that after predicting the \sep token, the model will decide what the next token will be. That means the model may predict track candidates in an arbitrary order.
Such predictions are repeated until all input tokens are in the output sequence and the last predicted token is the \sep token. This termination condition is not ideal when noise space points~\footnote{Noise space points are those created either from electronic noises or low-\pt particles (i.e. \pt $< 200$ MeV).} are presented as in real data. 

\begin{figure}[!htb]
    \centering
    \includegraphics[width=0.9\textwidth]{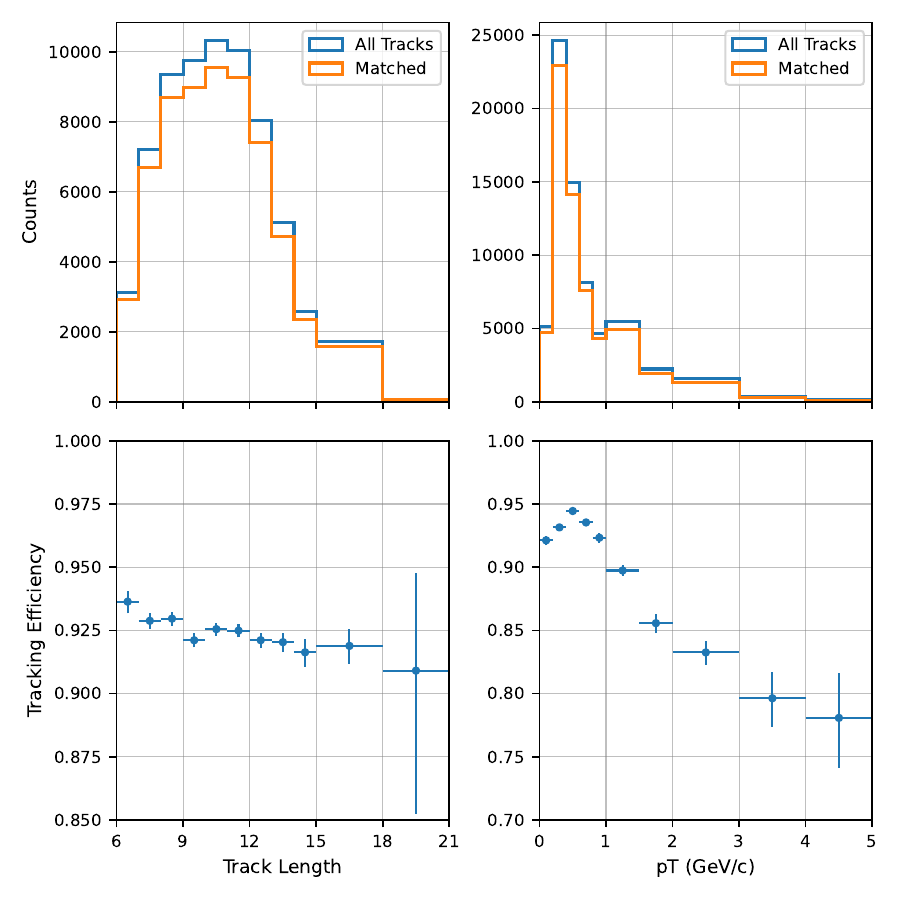}
    \caption{Top row: distribution of track length (left) and track \pt (right) in the test dataset. Bottom row: Tracking efficiency as a function of the track length (left) and particle transverse momentum (right).}
    \label{fig:results}
\end{figure}

The model performance is evaluated by the tracking reconstruction efficiency, defined as the fraction of particles that are matched to at least one reconstructed track. Reconstructed track candidates are  matched to particles if (1) 75\% of module hits in the reconstructed track are in the true particle track and (2) 75\% of module hits in the true particle track are in the reconstructed track. To assess our model, each track in our testing dataset was randomly paired with another track before being passed to the model. Fig.~\ref{fig:results} shows the model's performance with respect to track length and track \pt. 
The model performance is fairly stable against the track length, indicating the model can catch long-distance information. 
We note that tracking efficiency as a function of \pt resembles the distribution of particle \pt in our testing dataset, see figures on the top panel in Fig.~\ref{fig:results}. This suggests that an uniform sampling of particle \pt during training may make model performant in all \pt regions.  

We observed that a larger model size leads to better performance. We expect a larger dataset size to be beneficial to improve model performance. In our training dataset, we find several tokens represented in only a handful of samples. This adversely affects Word2Vec training of our initial embedding vectors as well as training of the model itself.

In smaller scale experiments focusing on the inner barrel detector region, we came up with physics-inspired embedding vectors for detector modules, such as the global coordinates, rotation matrix and pitch components, and geometric features of each detector module. The model performance resulting from this embedding scheme does not perform well compared to the more traditional Word2Vec implementation.

The model works fairly well with two tracks per input. but it remains to be studied whether the \modelname can scale effectively in a dense environment like the HL-LHC, where each event contains 10k particles resulting in 100k detector hits. 
This would form the event-level context window for the language model. 
This large context window may not pose a problem, as the leading LLM models already have substantial capacities. 
For instance, \textsc{GPT-4o} supports a 128k context window~\cite{gpt4o}, 
\textsc{Claude 3 Sonnet} extends up to 200k tokens~\cite{claude3sonnet}, 
and the \textsc{Gemini}-1.5 model can handle up to 1 million tokens in production~\cite{germin1p5}. 

\section{Conclusions}
\label{sec:conclusion}
We reformulated the particle tracking problem as a sequence to sequence problem and proposed a Tranformer-based track sorting algorithm to address it. This algorithm achieves good tracking reconstruction efficiency, even for low-\pt particles ($\pt < 1$ GeV). Our work leverages a language model-style architecture to tackle high-energy problems. Our study trained the language model from scratch. It remains to be studied open large language models can be fine-tuned for HEP problem-solving.

\section*{Data and Software availability}
Data can be found at Kaggle~\footnote{https://www.kaggle.com/c/trackml-particle-identification} and code is avaible at Github~\footnote{https://github.com/YashMelkani/Track-Sorting-Tutorial}.

\section*{Acknowledgements}
This research used resources of the National Energy Research Scientific Computing Center (NERSC), a U.S. Department of Energy Office of Science User Facility located at Lawrence Berkeley National Laboratory, operated under Contract No. DE-AC02-05CH11231.

\bibliography{main}

\providecommand{\href}[2]{#2}\begingroup\raggedright\begin{thebibliography}{10}

\bibitem{ATLAS:2008xda}
{\bf ATLAS} Collaboration, {\it {The ATLAS Experiment at the CERN Large Hadron
  Collider}},  {\em JINST} {\bf 3} (2008) S08003.

\bibitem{CMS:2008xjf}
{\bf CMS} Collaboration, {\it {The CMS Experiment at the CERN LHC}},  {\em
  JINST} {\bf 3} (2008) S08004.

\bibitem{ATLAS-TDR-25}
{\bf ATLAS} Collaboration, {\it {ATLAS Inner Tracker Strip Detector: Technical
  Design Report}}, .

\bibitem{ATLAS-TDR-30}
{\bf ATLAS} Collaboration, {\it {ATLAS Inner Tracker Pixel Detector: Technical
  Design Report}}, .

\bibitem{CMS:2017lum}
{\bf CMS} Collaboration, {\it {The Phase-2 Upgrade of the CMS Tracker}}, .

\bibitem{bert}
J.~Devlin, M.-W. Chang, K.~Lee, and K.~Toutanova, {\it Bert: Pre-training of
  deep bidirectional transformers for language understanding},
  \href{http://arxiv.org/abs/1810.04805}{{\tt arXiv:1810.04805}}.

\bibitem{gpt3}
T.~B. Brown, B.~Mann, N.~Ryder, M.~Subbiah, et~al., {\it Language models are
  few-shot learners},  \href{http://arxiv.org/abs/2005.14165}{{\tt
  arXiv:2005.14165}}.

\bibitem{llama:code}
B.~Rozière, J.~Gehring, F.~Gloeckle, S.~Sootla, et~al., {\it Code llama: Open
  foundation models for code},  \href{http://arxiv.org/abs/2308.12950}{{\tt
  arXiv:2308.12950}}.

\bibitem{grok1}
{\bf X-AI} Collaboration, {\it Grok-1},  2024.
\newblock \url{https://github.com/xai-org/grok-1}, accessed on July 30, 2024.

\bibitem{Birk:2024knn}
J.~Birk, A.~Hallin, and G.~Kasieczka, {\it {{OmniJet-}}\$\textbackslash
  alpha\$: {{The}} first cross-task foundation model for particle physics},
  \href{http://arxiv.org/abs/2403.05618}{{\tt arXiv:2403.05618}}.

\bibitem{vqvae}
A.~van~den Oord, O.~Vinyals, and K.~Kavukcuoglu, {\it Neural {{Discrete
  Representation Learning}}},  \href{http://arxiv.org/abs/1711.00937}{{\tt
  arXiv:1711.00937}}.

\bibitem{Heinrich:2024sbg}
L.~Heinrich, T.~Golling, M.~Kagan, S.~Klein, et~al., {\it Masked {{Particle
  Modeling}} on {{Sets}}: {{Towards Self-Supervised High Energy Physics
  Foundation Models}}},  \href{http://arxiv.org/abs/2401.13537}{{\tt
  arXiv:2401.13537}}.

\bibitem{Huang:2024voo}
A.~Huang, Y.~Melkani, P.~Calafiura, A.~Lazar, et~al., {\it {A Language Model
  for Particle Tracking}},  in {\em {Connecting The Dots 2023}}, 2, 2024.
\newblock \href{http://arxiv.org/abs/2402.10239}{{\tt arXiv:2402.10239}}.

\bibitem{bart}
M.~Lewis, Y.~Liu, N.~Goyal, M.~Ghazvininejad, et~al., {\it Bart: Denoising
  sequence-to-sequence pre-training for natural language generation,
  translation, and comprehension},  2019.

\bibitem{TrackMLAccuracy2019}
S.~Amrouche et~al., {\it {The Tracking Machine Learning challenge : Accuracy
  phase}},  \href{http://arxiv.org/abs/1904.06778}{{\tt arXiv:1904.06778}}.

\bibitem{Mikolov:2013ohu}
T.~Mikolov, K.~Chen, G.~Corrado, and J.~Dean, {\it Efficient {{Estimation}} of
  {{Word Representations}} in {{Vector Space}}},
  \href{http://arxiv.org/abs/1301.3781}{{\tt arXiv:1301.3781}}.

\bibitem{word2vec}
T.~Mikolov, I.~Sutskever, K.~Chen, G.~Corrado, et~al., {\it Distributed
  {{Representations}} of {{Words}} and {{Phrases}} and their
  {{Compositionality}}},  \href{http://arxiv.org/abs/1310.4546}{{\tt
  arXiv:1310.4546}}.

\bibitem{rehurek_lrec}
R.~{\v R}eh{\r u}{\v r}ek and P.~Sojka, {\it {Software Framework for Topic
  Modelling with Large Corpora}},  in {\em {Proceedings of the LREC 2010
  Workshop on New Challenges for NLP Frameworks}}, (Valletta, Malta),
  pp.~45--50, ELRA, May, 2010.
\newblock \url{http://is.muni.cz/publication/884893/en}.

\bibitem{transformer}
A.~{Vaswani}, N.~{Shazeer}, N.~{Parmar}, J.~{Uszkoreit}, et~al., {\it
  {Attention Is All You Need}},  {\em arXiv e-prints} (June, 2017)
  [\href{http://arxiv.org/abs/1706.03762}{{\tt arXiv:1706.03762}}].

\bibitem{adam}
D.~P. Kingma and J.~Ba, {\it Adam: A method for stochastic optimization},
  \href{http://arxiv.org/abs/1412.6980}{{\tt arXiv:1412.6980}}.

\bibitem{gpt4o}
{\bf Open-AI} Collaboration, {\it Gpt-4o},  2024.
\newblock \url{https://platform.openai.com/docs/models/gpt-4o}, accessed on
  July 30, 2024.

\bibitem{claude3sonnet}
{\bf Anthropic} Collaboration, {\it Claude 3 sonnet},  FEB, 2024.
\newblock \url{https://docs.anthropic.com/en/docs/about-claude/models},
  accessed on July 30, 2024.

\bibitem{germin1p5}
{\bf Google} Collaboration, {\it Gemini 1.5},  FEB, 2024.
\newblock
  \url{https://blog.google/technology/ai/google-gemini-next-generation-model-february-2024/#context-window},
  accessed on July 30, 2024.

\end{thebibliography}\endgroup
\bibliographystyle{jhep}

\end{document}